# Improved FedProx-TinyViT Model for Low-Voltage Distribution Area Topology Identification Under Smart Meter Voltage-Related Data

Jinlu Zhang[a]
[a] *State Grid Tianjin Binhai Electric Power Supply Company, Tianjin, 300450, Tianjin, China*

**Abstract.** This work presents an improved FedProx TinyViT model for low voltage distribution-area topology identification using smart-meter voltage related data under asynchronous sampling, missing meter packets, partial feeder records, and privacy constraints. The simulated dataset includes 126 transformer regions, 18,400 smart meters, 12 months of voltage, current, power factor, outage, and event records, and 42 controlled topology-change scenarios. Voltage correlation matrices, fluctuation response patches, missingness masks, and feeder event descriptors are constructed as topology-sensitive inputs. A TinyViT branch learns compact local relation patterns, while FedProx aggregation limits client drift across none independent transformer areas without centralizing raw customer curves. Compared with centralized graph learning, local only TinyViT, FedAvg TinyViT, and voltage correlation baselines, the proposed model reaches 94.6% topology identification accuracy, a branch user F1 of 0.927, a 1.4 day topology change detection delay, and a 28.5% reduction in cross-area false association. The findings suggest that proximal federated optimization and lightweight voltage relation patches can support privacy-preserving topology governance, while final topology-file updates still require operator review and field evidence.

## Introduction

To guarantee the distribution network's reliable operation, a low-voltage distribution-area topology must be chosen and regularly used in the actual field wiring [1]. Demand-side management, phase-imbalance control, loss analysis, outage monitoring, customer service response, and power-quality tracing [2] are all directly impacted by the architecture. Despite the fact that smart meters gather data on voltage, current, power factor, events, and outages, the customer information system might not be updated with later wire changes, meter replacements, feeder maintenance, or user transfers [3]. Voltage statistics [4] are helpful because users linked to the same transformer or branch frequently react similarly to load changes, voltage sags, switching operations, and local reactive-power correction. Even a good state-estimation or line-loss model will exhibit aberrant voltage behaviour on the incorrect feeder segment if the topology file is incorrect. Thus, under practical smart-meter restrictions, this study tackles topology identification as a data-driven relation-inference and governance challenge.

The low-voltage regions are not clean experimental circuits, which is the real issue. The same transformer district may include transient loads, rooftop solar inverters, charging piles, residential structures, and small business users. Interruptions in communication result in missed voltage sequences, and meter sampling periods are not always coordinated. Certain meters may exhibit anomalous readings following firmware changes, clock drifts, or replacement. The pertinence of federated learning [5] follows from the reluctance of grid firms to centralise the complete customer-level curves from all sectors. However, when one region has a high-density residential area and the other has a business or photovoltaic-dominated sector, the usual federated averaging becomes unstable. Successful topology inference should, according to smart-grid privacy studies [6], preserve user data near to the operational region and enable the model to share generic voltage-response information.

Lightweight visual and graph models have recently offered an effective solution to this issue. Users who react to the same feeder disturbance may be identified using local voltage fluctuation patches, and a voltage-correlation matrix can be thought of as a structured relation picture. For distribution-area models that must operate on edge servers instead than in massive cloud clusters, a compact representation akin to TinyViT is appropriate. In situations when the transformer region has distinct user distributions, federated optimisation with a proximal term can further minimise client drift. The concepts are combined in the proposed FedProx-TinyViT model, which updates global model parameters with a penalty for excessive local deviation, learns branch connection evidence using a lightweight attention backbone, and transforms voltage-related sequences into topology-sensitive patches.

The modelling route will be shown in the remaining sections of this study. In Section 2, federated lightweight grid analytics, voltage similarity mining, and smart-meter topology identification are reviewed. The FedProx-TinyViT learning process, voltage-matrix generation, and confidence-output design are demonstrated in Section 3. Using data figures and a single table, Section 4 assesses the simulated low-voltage dataset, topology correctness, ablation outcomes, topology-change detection, communication cost, and operational deployment. The main conclusion, weaknesses, and future research prospects are covered in Section 5.

## Related Work

### Smart Meter-Based Grid Topology Recognition

Voltage similarity, load response, outage coincidence, phase identification, and event association are typically used in smart-meter-based topology recognition. Because individuals in the same feeder section are typically impacted by correlated voltage variations after changes in the upstream load, voltage-similarity approaches are practical. When distributed solar systems have changing output or substantial load variety, voltage similarity alone might not be appropriate. A structural perspective is provided by depicting consumers, transformers, feeder nodes, and switching events as interconnected entities, an approach central to graph-based power-system learning [7]. Because the physical line file may be lacking and topology inference frequently needs to be completed before a full-field inspection is available, this structural view should nevertheless be able to handle questionable data in low-voltage locations.

### Lightweight Federated Models for Power Distribution Systems

The transformer component may be trained locally on voltage data to communicate just model updates without sending user data, making federated learning appropriate for smart-meter analysis. Although the distribution-area data are very non-independent, this privacy-preserving option is appropriate for customer-side metering records. Direct parameter averaging may produce an unstable global update since the variations in an industrial-edge area and a dense dwelling neighbourhood will differ significantly. The need for mechanisms that reduce client drift in federated learning on heterogeneous data is shown by Liu et al. [8], and the FedProx-style optimisation used here lowers this danger by adding a proximal constraint to local training. Because edge deployment cannot leverage a huge visual backbone, Lightweight Transformer Design is also required. The TinyViT-inspired modelling basis is provided by Wu et al. [9], allowing voltage-correlation patches to be processed as compact relation maps while maintaining enough local attention for branch detection.

### Data Quality and Privacy Issues in Low-Voltage Areas

Data quality is a core issue in topology identification rather than a secondary preprocessing detail. Missing meter packets, asynchronous clocks, abnormal voltage outliers, meter replacement, and transient user-side generation can distort pairwise voltage similarity. Missingness and noise should, as machine-learning data-quality studies [10] argue, be represented as part of the input instead of being silently discarded or over-smoothed. For this reason, the proposed topology model uses missingness masks and event descriptors together with voltage curves. The model also reports confidence, anomalous evidence, and review priority so that the operator can distinguish a low-quality data window from a likely physical topology change.

## Methodology

### Smart Meter Voltage Matrix Construction

Voltage-related data gathered by smart meters in a low-voltage distribution area serve as the input. Transformer IDs, available feeder branch labels, smart-meter voltage and current sequences, power-factor records, outage events, and communication-quality indicators are included for each region. The whole processing path from raw smart-meter records to topology confidence output is depicted in Figure 1. The workflow includes data alignment, anomalous sample marking, voltage-correlation matrix construction, fluctuation-response patch extraction, TinyViT feature learning, FedProx aggregation, topology decoding, and operator review output. Each training window covers seven consecutive days, each simulated transformer region contains 58 to 236 meters, and the sampling interval is 15 minutes.

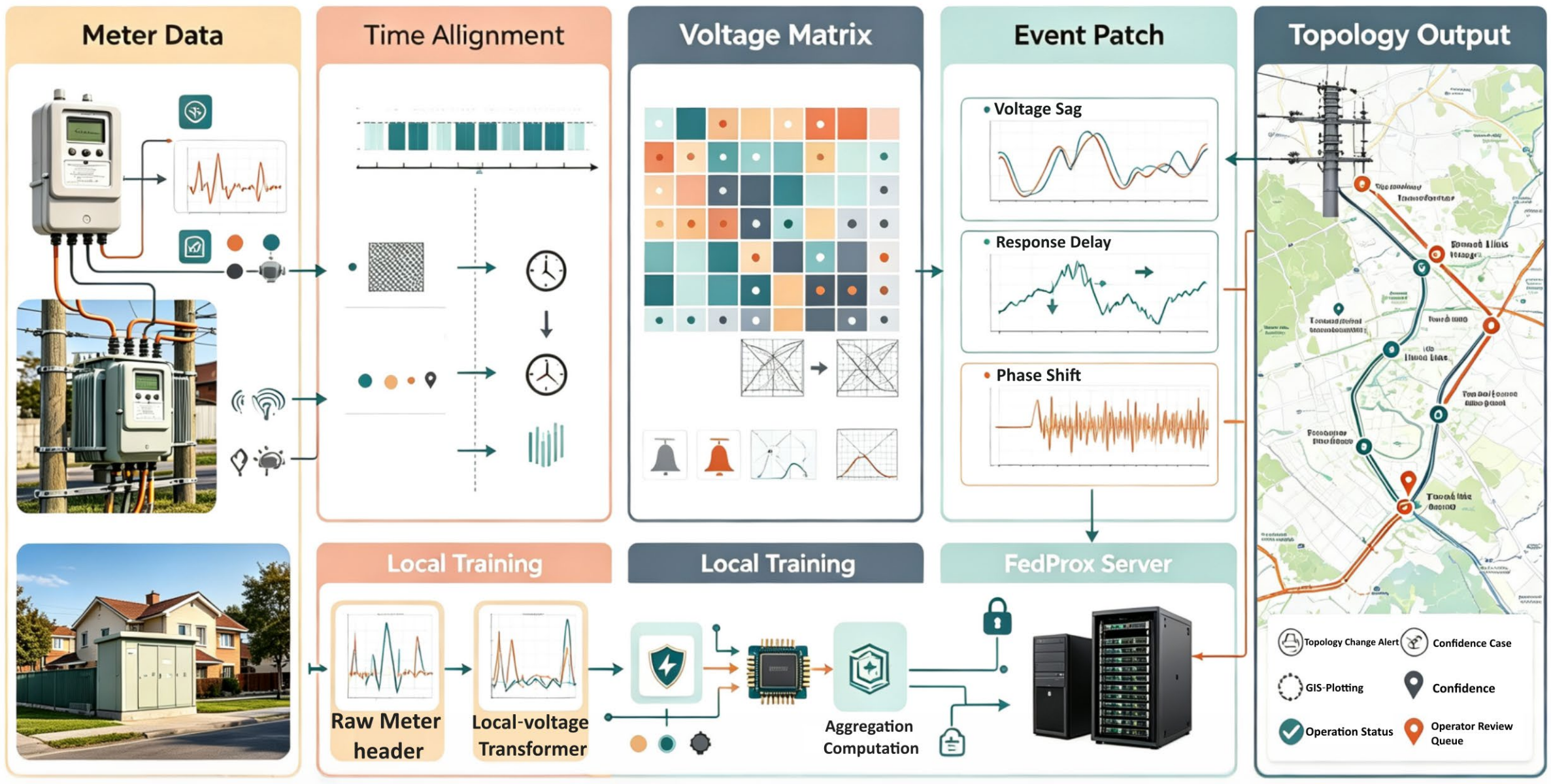

**Figure 1.** Smart meter voltage-topology identification workflow.

The voltage matrix is created by aligning meter sequences inside the same transformer area. Missing readings are not filled up at random; instead, they are left as a mask. This decision is required since the pattern of missingness itself may be untrustworthy evidence, and the communication breakdown may be restricted to a particular branch, meter box, or building. In order for the model to learn the general voltage response rather than absolute voltage levels, align the sequences and then normalise each user's data using their local mean and standard deviation.

$$v'_{i,t} = \frac{v_{i,t} - \mu_i}{\sigma_i + \epsilon} \quad \text{Eq.(1)}$$

Determine the user pair's voltage similarity. The correlation value is merely one channel of the relation patch and does not reveal the topology. While two distant users may seem identical during a transformer-wide voltage event, users in the same branch exhibit a strong association during a load variation. Consequently, reaction latency, missingness, and event evidence are connected with the later model.

$$C_{ij} = \rho(v'_i, v'_j) e^{-\eta \Delta\tau} \quad \text{Eq.(2)}$$

Around a voltage disruption, extract a Fluctuation-Response Patch. In this paper, a disturbance is defined as an area-level voltage derivative greater than the 92nd percentile or more than 18% of meters showing a brief sag within the same hour. The only purpose of these thresholds is to choose the informative window. The model must learn how users react to a variety of stimuli, and topology is not immediately provided.

$$R_{i,t} = [\Delta v_{i,t}, \Delta p_{i,t}, q_{i,t}, m_{i,t}] \quad \text{Eq.(3)}$$

Pairwise relation maps and event-response maps are combined in the topology-sensitive input patch. For the TinyViT branch, each patch is scaled to 64 by 64. The empty rows are hidden when there are fewer users in a certain location. Before patch construction, users are arranged by meter box, phase label, and historical feeder suggestion when there are more users. This grouping preserves the useful electrical linkages while lowering the cost of pairwise comparison.

### FedProx-TinyViT Topology Learning

The structure of the model is shown in Figure 2. Voltage-correlation matrices, fluctuation-response patches, missingness masks, and feeder-event descriptors are placed on the input side. A TinyViT feature extractor with window attention and lightweight feed-forward blocks learns compact relation features from these topology-sensitive patches. The topology decoder predicts transformer-user relation, branch-user relation, topology-change probability, and confidence state. Raw smart-meter curves remain inside the local distribution area, while the federated server aggregates area-side model updates through the FedProx objective.

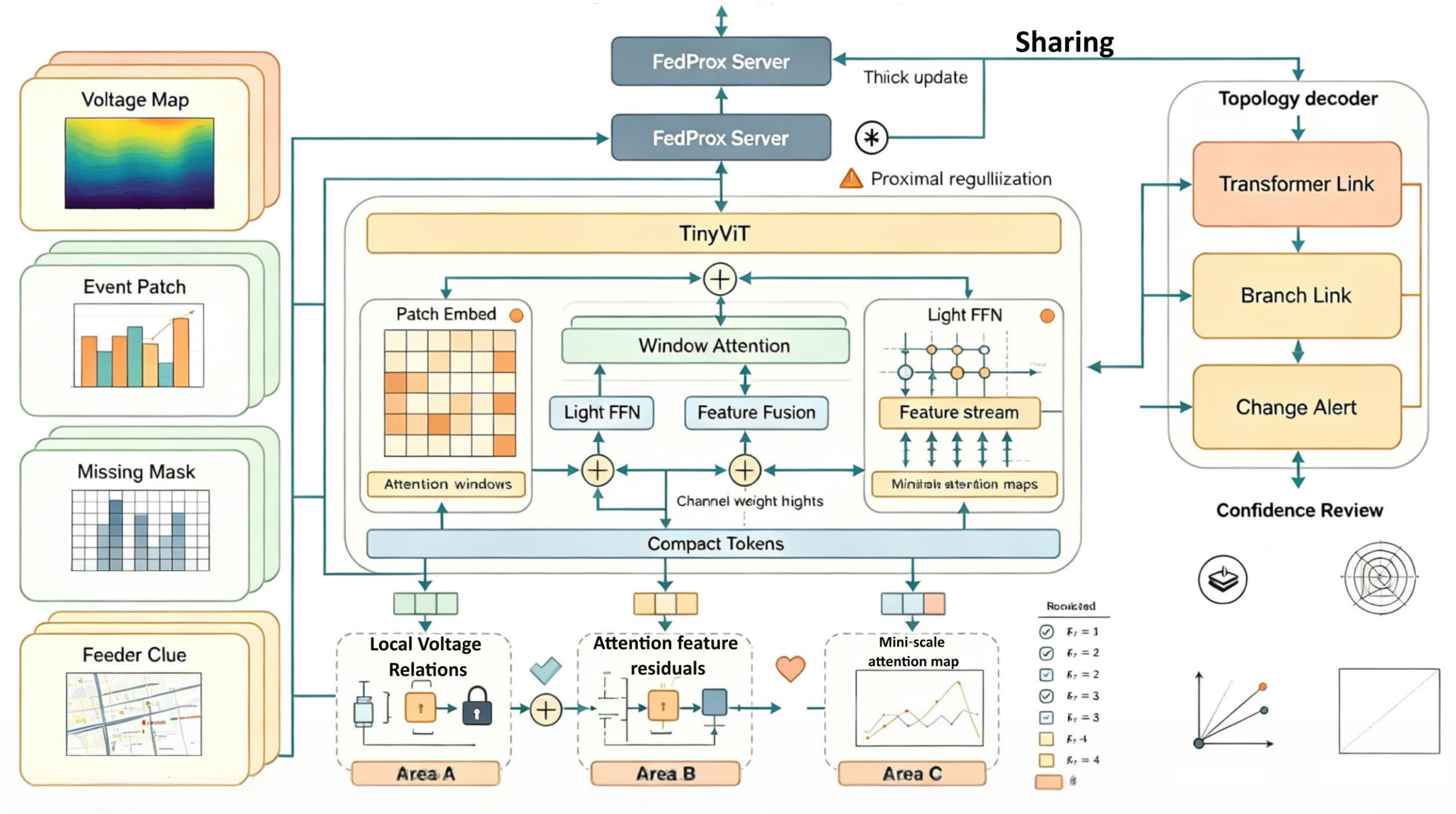

**Figure 2.** FedProx-TinyViT topology learning structure.

The TinyViT branch maps the voltage relation patch into a compact feature representation. Window attention is used because local groups of users often share stronger electrical relations than all users in the area. This design also reduces computation when transformer areas contain many meters.

$$z_l = T_l(P_l, M_l, E_l) \quad \text{Eq.(4)}$$

Local training minimizes topology classification loss, branch association loss, topology-change loss, and a proximal penalty that keeps the client model close to the global model. The proximal term is important for non-independent transformer areas because it reduces the chance that an unusual photovoltaic-heavy or commercial client pulls the global model away from shared voltage-response knowledge. In engineering terms, the term regularizes local learning so that area-specific behavior is captured without allowing one special area to dominate the topology model.

$$L_k = L_t + \lambda_b L_b + \lambda_c L_c + \frac{\mu}{2}\left\|\theta_k - \theta_g\right\|^2 \quad \text{Eq.(5)}$$

The global model is updated by weighted aggregation of local model parameters. The weight depends on the number of valid training windows in each area rather than only the number of meters, because an area with many missing meter packets should not dominate the global update.

$$\theta_g = \sum_k \alpha_k \theta_k \quad \text{Eq.(6)}$$

The final topology score is calculated from relation probability, branch probability, topology-change probability, and uncertainty. The score is used to form a review queue for operators. A relation with high probability and low uncertainty can be accepted automatically in the simulated workflow, while a relation with high uncertainty is sent to field verification or meter-file inspection.

$$S_{ij} = \sigma(w^{\top} y_{ij}) \quad \text{Eq.(7)}$$

70% of the transformer areas are utilised for local model training, 10% for validation, and 20% for testing. For the purpose of evaluating cross-area transfer, commercial and photovoltaic-heavy regions are purposefully kept in the test set. Each local client trains for five epochs throughout each of the server's 80 communication cycles. TinyViT has an embedding dimension of 128, a local batch size of 32, and a proximal coefficient of 0.01 following validation. The model can still perform topology-change discovery in noisy meter data with the parameters while being small enough for the distribution-edge server.

$$Q = \mathrm{sort}(S, u, \mathrm{type}\,, \mathrm{area}\,) \quad \text{Eq.(8)}$$

The output queue lists suspicious transformer-user links, uncertain branch-user links, and potential topology-change events. Each record contains the affected area, candidate relation, confidence score, voltage evidence window, missingness state, and suggested review action. This output is necessary because topology identification is not only a classification task; it is a data-governance workflow for distribution operators.

## Experiments and Discussion

### Dataset and Voltage Relation Characteristics

The simulated dataset has 126 low-voltage distribution areas, 18,400 smart meters, 12 months of voltage-related records, and 42 controlled topology-change scenarios. The proportion of residential areas for meters is 48.6%; mixed-use commercial areas are 28.9%; photovoltaic-influenced areas are 12.7%; and industrial-edge users account for 9.8%. Figure 3 is the dataset composition and voltage-event distribution. Figure 3(a) is the area-type distribution, and it shows that not all voltage sequences are in sync with the load rhythm of that area. Residential users are in the evening, commercial users are in the daytime, and photovoltaic users may produce a reverse-voltage rise at noon. Figure 3(b) shows meter-count intervals in transformer areas. Small areas with fewer than 80 metres are easier to identify because the pairwise voltage relationships are less dense; large areas above 180 metres have many candidate relationships and thus increase the risk of false association. An interval view can also be used to determine whether the model has sufficient examples of dense transformer districts, as branch-user inference will be significantly more difficult if the relation matrix contains many visually similar blocks. Figure 3(c) shows the voltage-event categories: sag, rise, fluctuation burst, outage coincidence and phase imbalance. Rather than a single Pearson correlation map, multiple voltage-related signals are needed in topology identification [11]. Therefore, the figure is more than a sample balance. Determine whether the experimental data have enough operating events to show the hidden wiring relationship. If the dataset only contained smooth daily voltage curves, most branch relations would be undetectable, and the model would learn user-load similarity rather than electrical connections.

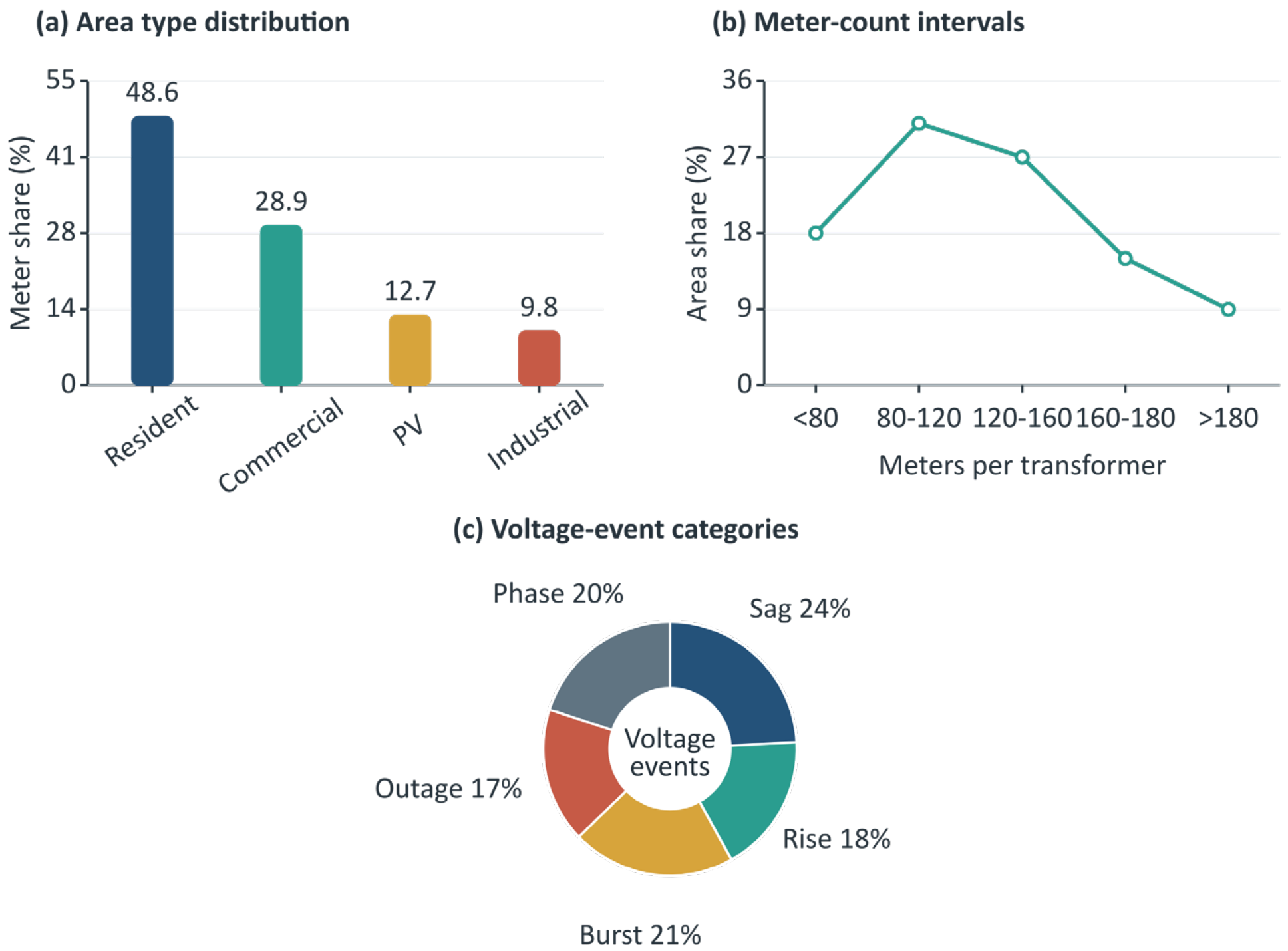

**Figure 3.** Dataset composition and voltage-event distribution. (a) Area type distribution. (b) Meter-count intervals. (c) Voltage-event categories.

Data-quality patterns are summarized in Figure 4. Figure 4(a) reports missing-packet levels, Figure 4(b) reports sampling-offset intervals, and Figure 4(c) reports abnormal readings and topology-change evidence. Missingness, sampling offsets, meter replacement, firmware updates, flat voltage, spike events, feeder maintenance, phase corrections, and meter-file mismatches can all reduce voltage similarity. However, they do not have the same operational meaning. The revised interpretation [12] therefore emphasizes that missingness masks and event descriptors are used to distinguish communication or meter-quality faults from physical topology changes.

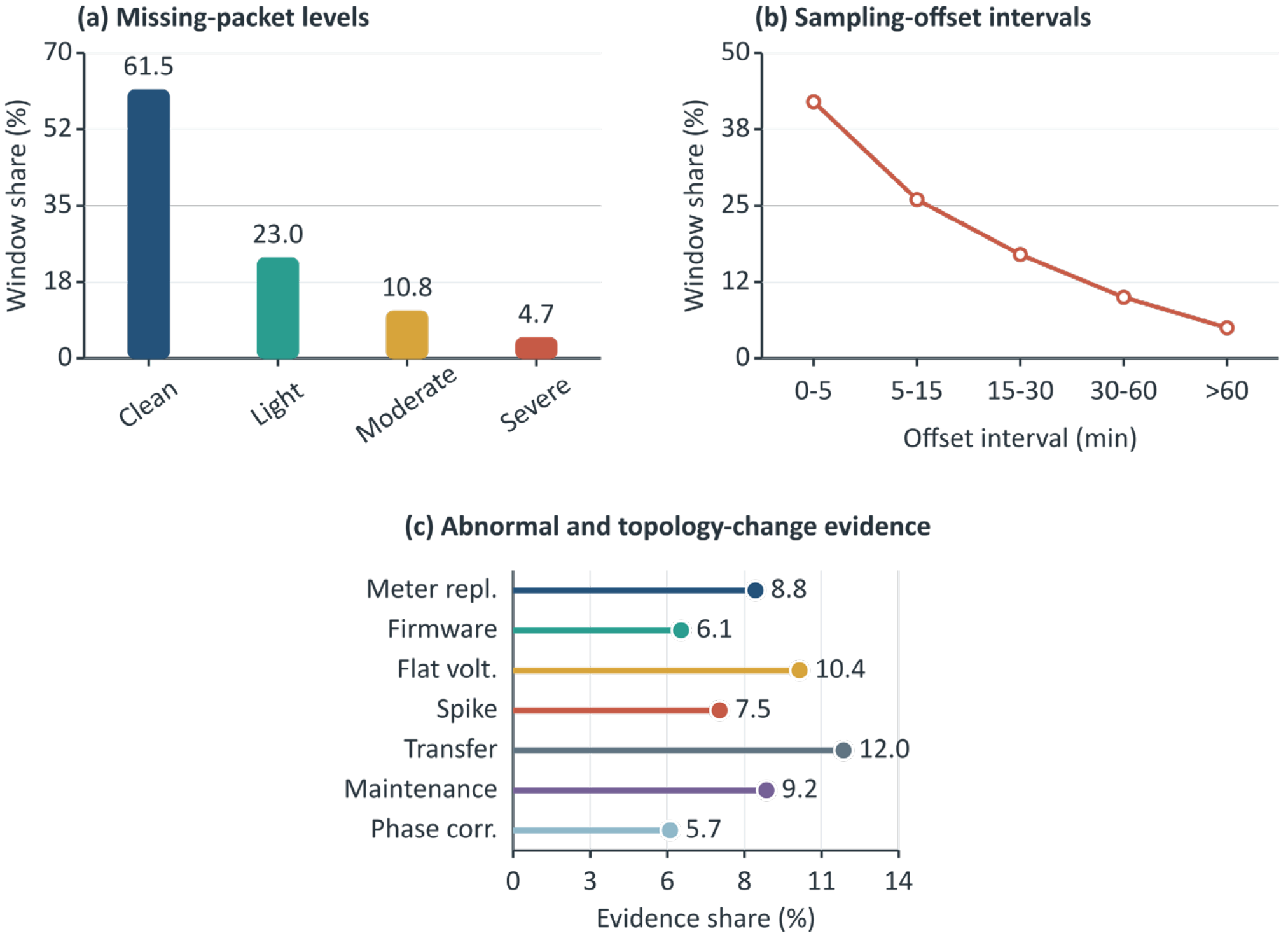

**Figure 4.** Data-quality and topology-change evidence. (a) Missing-packet levels. (b) Sampling-offset intervals. (c) Abnormal and topology-change evidence.

A dataset of Transformers has been divided by Transformer area, not randomly. This avoids the same area appearing in both training and testing with almost identical voltage patterns. Federated evaluation also uses a client-side setting; each transformer area trains locally and uploads model updates, but the server never receives raw meter curves. This Design investigates whether the model has learned transferable voltage-response knowledge across different user groups. There are 17 topology-change events and 8 photovoltaic influence areas above 25% of the midday load in the test set. These areas have been intentionally preserved because they are cases where the general similarity method is most likely to fail. This split strategy is supported by graph-based power-system learning [13], since structural inference needs to be evaluated on unseen graph configurations, not just repeatedly sampled ones. A split will not allow the model to learn a fixed user order in the correlation matrix. If a random-window split were employed, the same meter group could appear in both the training and testing sets; thus, the model might appear accurate simply because it had previously seen that same voltage block. The area-level separation is relatively coarse; thus, it is less representative of actual operating tasks and may fail to detect topologies in new or recently modified transformer areas.

Voltage relation maps also show why TinyViT-style patch modelling is advantageous. A clean residential Area has been used to group same-branch users in the correlation matrix. In a photovoltaic-heavy Area, the block shape is weaker due to local fluctuations in inverter output. User sequences in a commercial area may exhibit similar daytime patterns even if they belong to different branches. A relation-image model can learn these block and non-block patterns; a scalar similarity rule generally cannot. Similar aggregate curves can conceal different device or user structures, as research on building-load datasets [14] also shows. Therefore, the experiment will also include transformer-user accuracy, branch-user F1, topology-change detection delay and false association rate. A model that performs well in the transformer-user relationship may still be poor at branch-level identification. A model with a high branch F1 may still fail after a sudden topology change following a maintenance operation. The relation-map view can be used to find the cause of the error. If the wrong relationship only occurs at the peak output of photovoltaics, it can be designated as generation-sensitive. If the wrong relation occurs during all disturbance windows, it may be a real topology-file mismatch. The model will likely be used to support topology governance, so it should not be a "black box" that lacks an operational explanation.

### Topology Identification Performance

The main quantitative results are shown in Table 1. The proposed FedProx-TinyViT model reaches 94.6% transformer-user accuracy, a branch-user F1 of 0.927, a 1.4-day topology-change delay, and 28.5% lower false association than the strongest baseline. Centralized graph learning is competitive when all data are available, but it is not privacy-preserving and is sensitive to missing-packet bursts. Local TinyViT avoids centralized curves but overfits to individual areas, while FedAvg-TinyViT can drift under non-independent area distributions. By reducing this drift, the proximal term also regularizes branch-relation learning [15]. Table 1 therefore supports an operational tradeoff rather than a simple top-line accuracy claim.

**Table 1**. Quantitative topology identification results under smart-meter voltage-related data.

| Model | Topo. Acc. (%) | Branch F1 | Delay (days) | False Assoc. (%) |
|---|---|---|---|---|
| Voltage Corr. | 84.2 | 0.781 | 5.6 | 12.8 |
| Graph Learning | 90.5 | 0.861 | 3.8 | 9.7 |
| Local TinyViT | 88.9 | 0.842 | 4.2 | 10.9 |
| FedAvg-TinyViT | 92.1 | 0.893 | 2.6 | 8.3 |
| FedProx-TinyViT | 94.6 | 0.927 | 1.4 | 5.9 |

Figure 5 shows topology identification results in greater detail. Figure 5(a) reports transformer-user accuracy across residential, commercial, photovoltaic, and industrial-edge areas. Figure 5(b) reports branch-user F1 and cross-area false association. The proposed model remains stable in residential and mixed-use areas and loses less accuracy than competing baselines in photovoltaic-heavy areas, where local generation can distort voltage similarity. Branch-user F1 and false association are emphasized because an incorrect branch relation can damage outage tracing, phase checking, and loss accounting even when transformer-level accuracy appears high. The use of a compact attention backbone for relation-block recognition is supported by lightweight Transformer research [16].

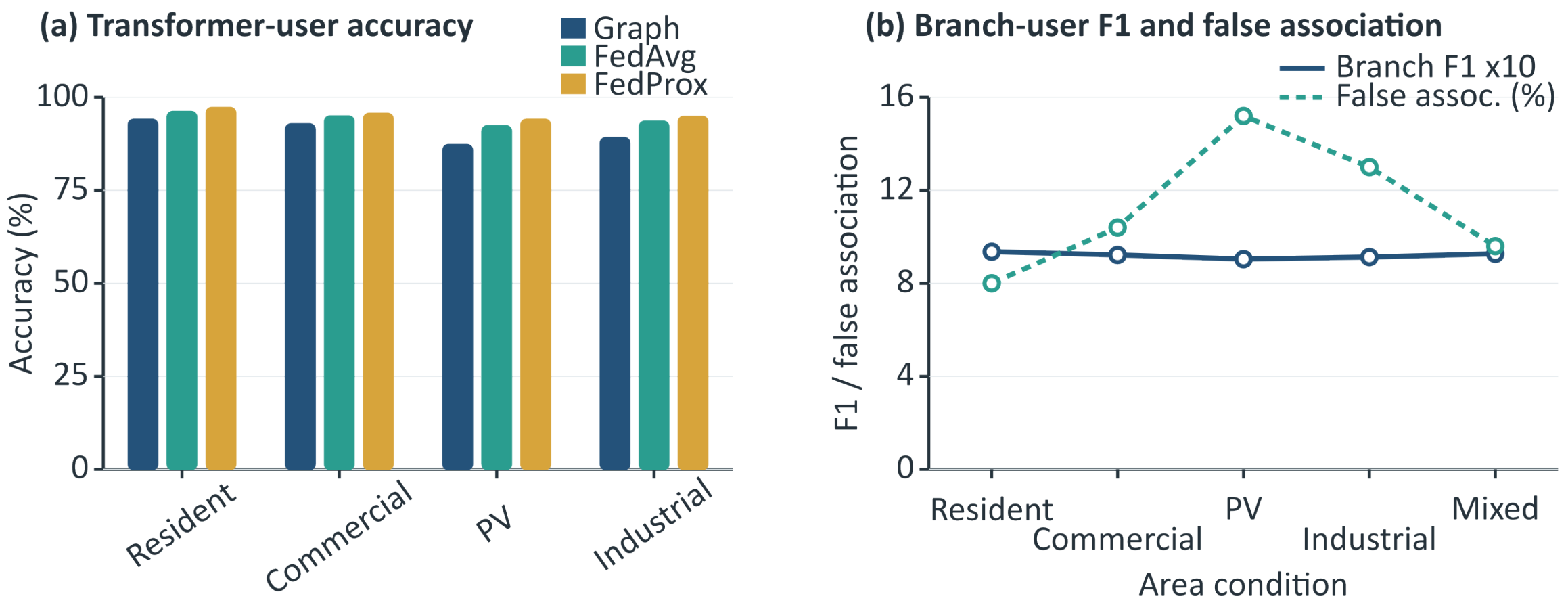

**Figure 5.** Topology identification accuracy and relation reliability. (a) Transformer-user accuracy. (b) Branch-user F1 and false association.

Figure 6 shows topology-change detection behavior. Figure 6(a) reports detection delay, Figure 6(b) reports event-level precision and recall, and Figure 6(c) reports confidence distribution. User transfer and feeder maintenance events are detected faster than meter-file mismatch events because physical wiring changes alter voltage response more directly. Meter-file mismatch may require several informative voltage events before the evidence becomes strong. Confidence is also higher in clean areas and lower under severe missingness or photovoltaic fluctuation. Uncertain topology relations should be sent to field verification rather than accepted automatically [17], which makes this behavior necessary.

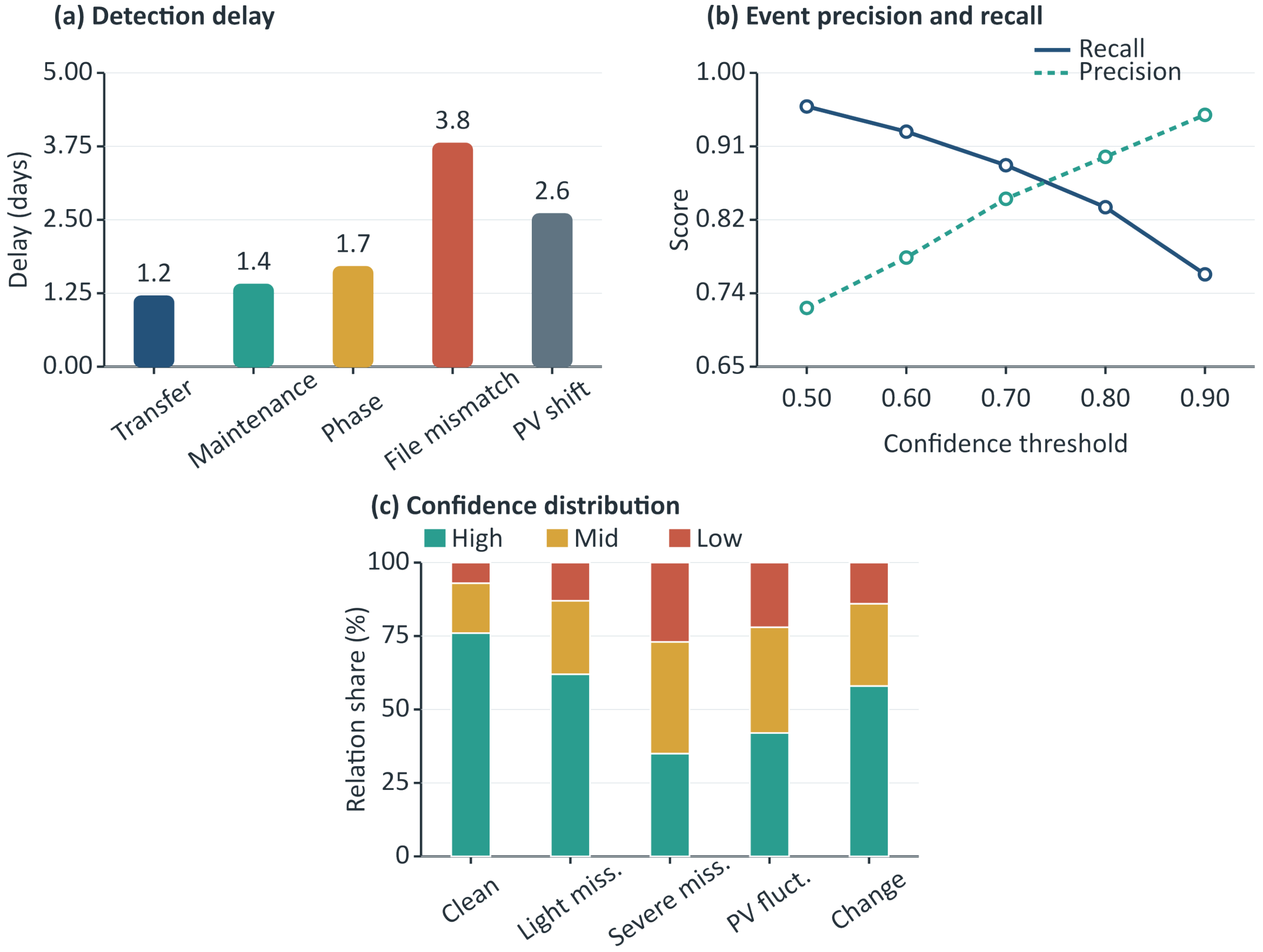

**Figure 6.** Topology-change detection behavior. (a) Detection delay. (b) Event precision and recall. (c) Confidence distribution.

Ablation results show how much the modules are increased. Removing missingness masks reduces the branch-user F1 by 2.8 points because the model considers communication gaps as normal voltage behaviour. Removing event descriptors reduces topology-change recall by 4.1 points because the model no longer has information on outage coincidence and voltage sag timing. Omitting the proximal term will increase the false association for non-independent clients, especially in photovoltaic-heavy and commercially-oriented Areas. Removing TinyViT window attention reduces transformer-user accuracy by 3.2 points because the local relation blocks are less effective. Based on the above results, both electrical-relation features and federated-optimisation stability are required by the Design Principle for Topology Identification [18]. Ablation experiments show that all modules are necessary for different parts of the task. Missingness generally enhances the reliability of poor-communication environments. Event descriptors generally enhance the timeliness of topology-change discovery. Proximal term extends cross-area generalisation, and window attention learns relation-pattern recognition in a single topology patch. This division of functions is convenient for engineering deployment; thus, an operator can choose to reinforce only the module with a change in local data conditions. For example, if an area has stable communication but frequent fluctuations in photovoltaic power, event descriptors and reactive-power features should be used. If a region has many packet losses, first perform missingness modelling and adjust confidence thresholds. If several clients have significantly different user structures, the proximal coefficient can be increased to prevent local drift of the global model. Error cases are also informative: wrong branch assignments occur when neighbouring branches serve similar commercial loads and experience almost identical voltage changes, and false topology-change alarms may arise after meter replacement due to a shifted timestamp and a slightly different voltage calibration of the new meter. Ambiguity also occurs in the user-side photovoltaic generation. This ambiguity can be reduced by adding voltage along with current and reactive-power features, as distribution-network data analytics [19] shows; furthermore, area-side clients should, as federated IoT research [20] suggests, keep the local uncertainty instead of following a single global pattern. Another error group occurs in a very small transformer area, and one abnormal meter can distort the sparse relation matrix. Topology confidence should be understood in light of the outage record and the installation archive.

**Communication Cost and Deployment Review**

Federated deployment is judged by communication rounds, client drift, edge inference time, and operator review workload. Figure 7(a) compares convergence curves under FedAvg and FedProx. FedAvg performs well early but oscillates when photovoltaic-heavy clients join aggregation. FedProx converges more slowly during early rounds, but its validation curve becomes more stable after repeated communication. Figure 7(b) combines client-drift magnitude and communication volume. The proximal penalty reduces area-side parameter deviation by 34.7%, and the compressed TinyViT update remains below 18.6 MB. These results show that deployment cost is not only a bandwidth issue but also a stability issue. Combining local-data retention with efficient model-update exchange is supported in both federated smart-grid learning [21] and privacy-preserving learning studies [22].

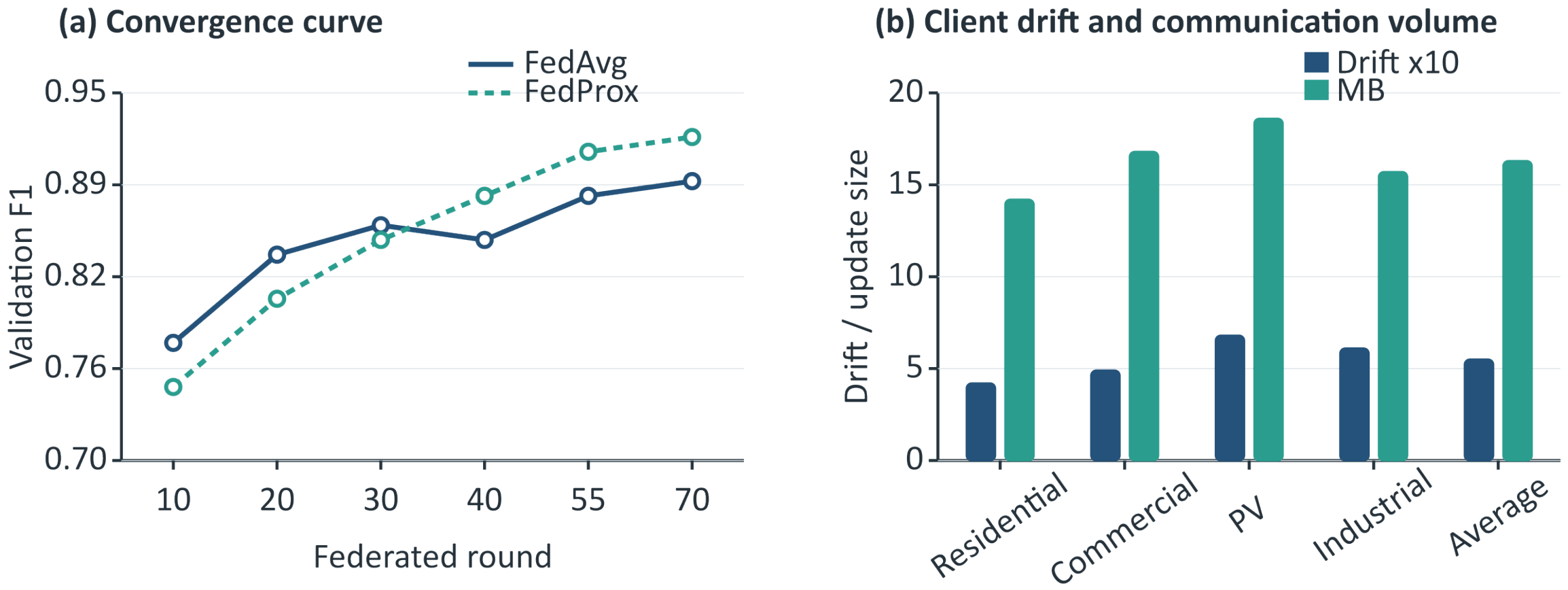

**Figure 7.** Federated training and communication behavior. (a) Convergence curve. (b) Client drift and communication volume.

The operating review results are shown in Figure 8. Figure 8(a) reports the relation-review queue, including confirmed relations, suspected branch relations, topology-change candidates, and meter-file mismatches. Figure 8(b) reports manual workload reduction, showing that the proposed model reduces manual relationship verification by 41.2% compared with correlation-based screening. Each review item contains the candidate user pair, affected transformer area, key voltage window, missingness state, event type, confidence reason, and suggested action. Connecting model output to human review is supported by edge AI deployment research [23], whereas smart-meter consumption analysis [24] warns that raw model scores should be interpreted together with operational data. The revised text also explains that rejected relations can become hard-negative examples for the next local training round.

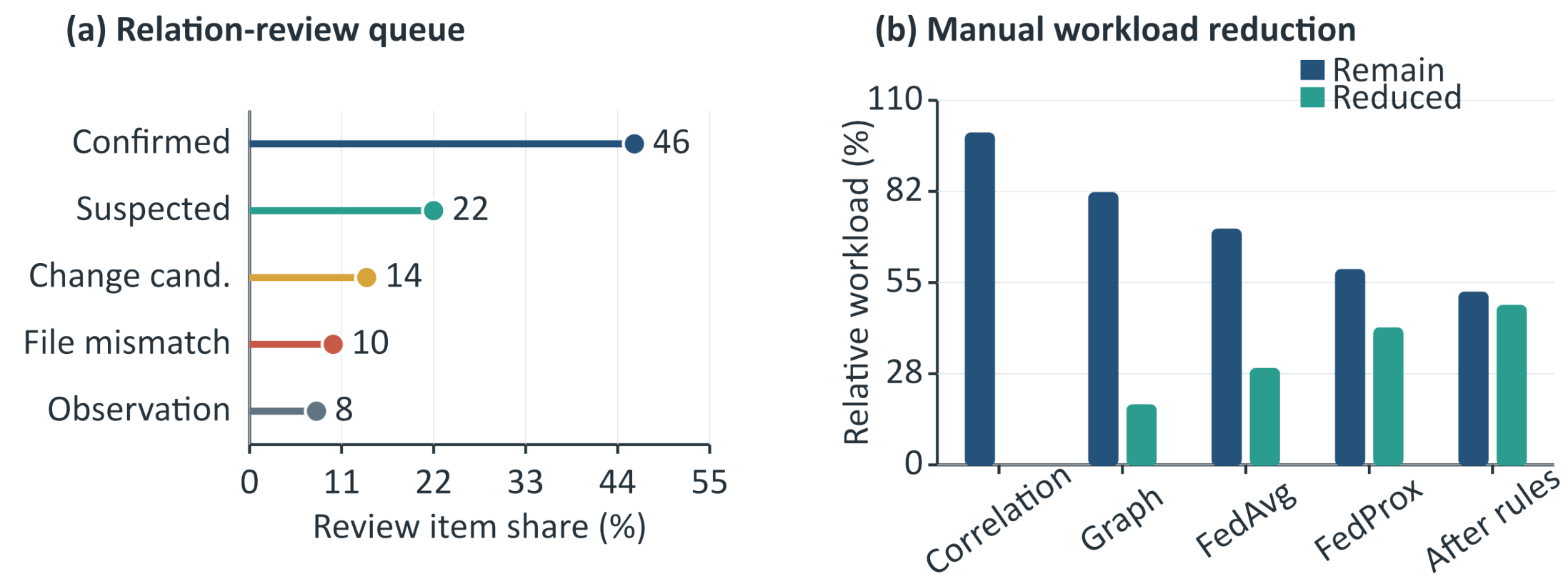

**Figure 8.** Operator review and topology governance output. (a) Relation-review queue. (b) Manual workload reduction.

Deployment also requires careful data governance. The topology inference system should not overwrite the grid archive silently when evidence is uncertain. In the simulated workflow, automatic acceptance is allowed only when confidence exceeds 0.91 and uncertainty is below 0.08 for three consecutive informative windows. Relations between 0.72 and 0.91 are sent to file checking, and lower-confidence relations are kept for observation. This conservative design is supported by smart-grid explainability work [25]: operators must know whether a recommendation is caused by voltage fluctuation, outage coincidence, missingness, or federated confidence. Field records and data-driven evidence need to be reconciled, as topology-identification studies [26] likewise show.

The three functions of the model are as follows. First, improve topology-file cleaning by identifying users whose voltage behaviour does not match that of the recorded transformer. Second, it supports branch-level phase and feeder checks by dividing into uncertain user groups. Third, it reduces the lag between a physical topology change and a data-system update. In the simulated 12-month period, among 42 topology-change scenarios, 36 were identified within two days and 5 within five days. One case has not been covered because the affected meters had a communication failure during the key event. As shown above, the model can enhance data governance but does not eliminate the requirement for field verification. Another relation cue for this review stage can also be provided through phase labels by customer phase-identification research [27]. This way, the voltage event can be either natural or signaled by a change in switching operation. Topology evidence should be linked to downstream operational use rather than being saved as a separate model result, as applied energy analytics [28] proposes. If the area of flat voltage and small current changes is prolonged, topology inference will be unreliable. Why local generation may alter the voltage characteristics in a manner similar to topology change is shown by smart-grid pricing and distributed-generation research [29]. In the future, integrate feeder terminal data, transformer low-voltage outlet measurements and phase identification records to enhance the topology evidence. A reference for combining model confidence with physical consistency checks is also offered by probabilistic topology identification [30]. The first problem in terms of deployment is that all topology relationships cannot be automatically resolved. The system can reduce the quantity of user records to be inspected by an operator from thousands to a small number of suspicious relationships that have visible evidence. Topology governance will be transformed from passive archive-correction into an active data-quality process. It also supports multiple checks after maintenance, meter replacement and seasonal photovoltaic variations, and the same transformer area may need several rounds of evidence before the final topology update is approved.

## Conclusion

This research proposes an enhanced FedProx-TinyViT model for topology identification in low-voltage distribution areas based on voltage-related data from smart meters. Create topology-sensitive patches utilising feeder events, voltage sequences, fluctuation responses, and missingness masks; use a compact TinyViT backbone to gather evidence of local relations; and use FedProx aggregation to reduce client drift in diverse transformer areas. The simulated experiment revealed a 1.4-day topology-change detection delay, 0.927 branch-user F1, and 94.6% topology correctness. According to the findings, if operators examine uncertain relations before adding them to the archive, voltage-relation pictures and privacy-preserving federated learning can allow realistic topology governance.

The studies still have shortcomings. Real low-voltage locations could have more serious meter-clock issues, mixed public and private metering devices, unrecorded user-side solar installations, delayed field-maintenance reports, feeder-map ambiguity, and archive anomalies since the dataset is simulated. a model that reacts to changes in voltage. Branch-level inference will be unreliable if key meters lose communication during a topology change or if all users in an area have received almost flat voltage for a long time. Thus, field-verification criteria, repeated evidence, and local operating restrictions should limit automated topology-file changes.

The model will be applied to actual transformer area data, physical feeder maps, phase identification records, and low-voltage outlet measurements in further study. To update topology files without disclosing user-level curves for distribution operators, secure aggregation, differential privacy, adaptive client selection, and live topology-change learning should also be used. Connecting topology inference with loss analysis, phase imbalance correction, outage tracing, and distributed photovoltaic hosting-capacity evaluation is another interesting approach that might transform the model into a module for a large-scale low-voltage digital-grid management workflow.